\documentclass[letterpaper, 10 pt, conference]{ieeeconf}  
\IEEEoverridecommandlockouts                              

\usepackage[utf8]{inputenc} 
\usepackage{graphics}   
\usepackage{amssymb, amsmath}   

\usepackage{amsthm}     
\usepackage{microtype}  
\usepackage{epstopdf}   
\usepackage{url}    
\usepackage[noadjust]{cite} 
\usepackage{overpic}    
\usepackage[linesnumbered,ruled,vlined]{algorithm2e}    
\usepackage[dvipsnames]{xcolor} 
\usepackage{wrapfig}
\usepackage[font=footnotesize,labelfont=bf]{caption}
\usepackage{multirow}   
\usepackage{footnote}   
\usepackage{soul}   
\usepackage{tikz}   
\let\labelindent\relax
\usepackage[inline]{enumitem}
\usetikzlibrary{tikzmark}
\usepackage{svg}
\usepackage{siunitx}
\usepackage[textsize=scriptsize]{todonotes}
\usepackage{comment}

\definecolor{lightyellow}{RGB}{240,226,182}

\DeclareMathOperator*{\argmax}{arg\,max}

\makeatletter
\let\NAT@parse\undefined
\makeatother
\definecolor{darkblue}{rgb}{0.0, 0.0, 0.55}
\usepackage[bookmarks=true, hidelinks]{hyperref}    
\hypersetup{
     colorlinks   = true,
     linkcolor= darkblue,
     citecolor    = darkblue,
     urlcolor=black
}
\SetKwProg{Fn}{Function}{}{}
\SetKwComment{Comment}{$\triangleright$\ }{}

\def\prob{\textsc{REMP}\xspace}
\def\hbfs{\textsc{HBFS}\xspace}
\def\pmmr{\textsc{PMMR}\xspace}

\font\titlefont=ptmb at 14.5pt
\title{\titlefont
Toward Optimal Tabletop Rearrangement with Multiple Manipulation Primitives
}
\author{Baichuan Huang\quad Xujia Zhang\quad Jingjin Yu     
\thanks{B. Huang and J. Yu are with the Department of 
Computer Science, Rutgers, the State University of New Jersey, Piscataway, NJ, USA. 
Emails: {\tt\small \{baichuan.huang, jingjin.yu\}@rutgers.edu}.
X. Zhang is with Southern University of Science and Technology, China. This work was done when X. Zhang was an intern at Rutgers University. 
}%
}

\begin{document}

\maketitle
\thispagestyle{empty}
\pagestyle{empty}

\begin{abstract} 
In practice, many types of manipulation actions (e.g., pick-n-place and push) are needed to accomplish real-world manipulation tasks. Yet, limited research exists that explores the synergistic integration of different manipulation actions for optimally solving long-horizon task-and-motion planning problems. 
In this study, we propose and investigate planning high-quality action sequences for solving long-horizon tabletop rearrangement tasks in which multiple manipulation primitives are required. Denoting the problem \emph{rearrangement with multiple manipulation primitives} (\prob), we develop two algorithms, \emph{hierarchical best-first search} (\hbfs) and \emph{parallel Monte Carlo tree search for multi-primitive rearrangement} (\pmmr) toward optimally resolving the challenge. Extensive simulation and real robot experiments demonstrate that both methods effectively tackle \prob, with \hbfs excelling in planning speed and \pmmr producing human-like, high-quality solutions with a nearly $100\%$ success rate. 
Source code and supplementary materials will be available at \href{https://github.com/arc-l/remp}{\texttt{\textcolor{blue}{https://github.com/arc-l/remp}}}. 
\end{abstract}

\section{INTRODUCTION}
\label{sec:intro}

Real-world manipulation tasks, e.g., rearranging a messy tabletop or furniture in the house, often require multiple manipulating primitives (e.g., pick-n-place, pushing, toppling, etc.) to accomplish. 
When rearranging small/light objects, e.g., a cell phone on a table or a small chair in a room, it is convenient to do a \emph{pick-n-place}, i.e., to pick up the object, lift it above other objects, move it across the space to above its destination on the table, and then place it. 
On the other hand, for handling large/heavy objects, e.g., a thick book or a heavy couch, \emph{pushing} or \emph{dragging} close to the space's surface is more commonly adopted, executed with added caution. In this case, planning the object's motion trajectory must consider avoiding colliding with other objects more carefully. 
Solving such long-horizon task-and-motion planning tasks efficiently and optimally is highly challenging, as it involves not only an extended horizon but also selecting among multiple types of manipulation primitives at each step, both of which add to the combinatorial explosion of the search space. 
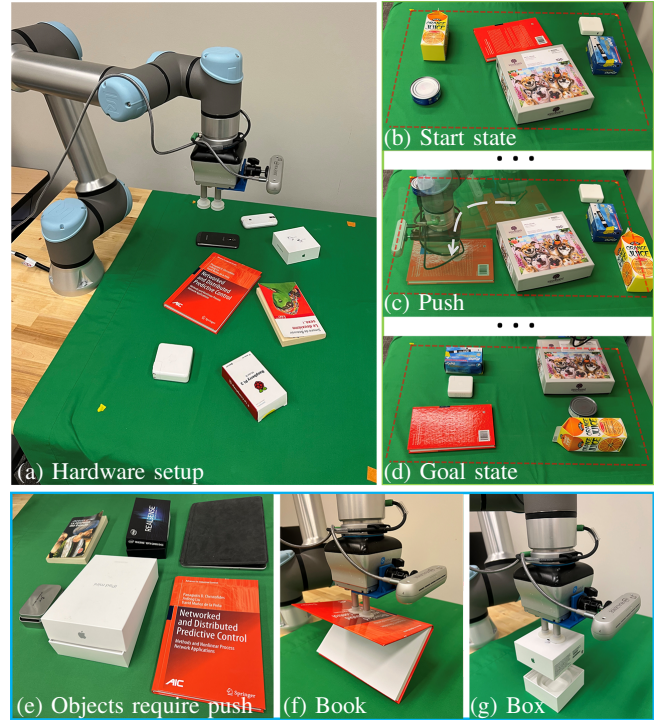
\begin{figure}[t!]
\vspace{1.5mm}
    \centering
    \begin{overpic}[width=0.99\linewidth]{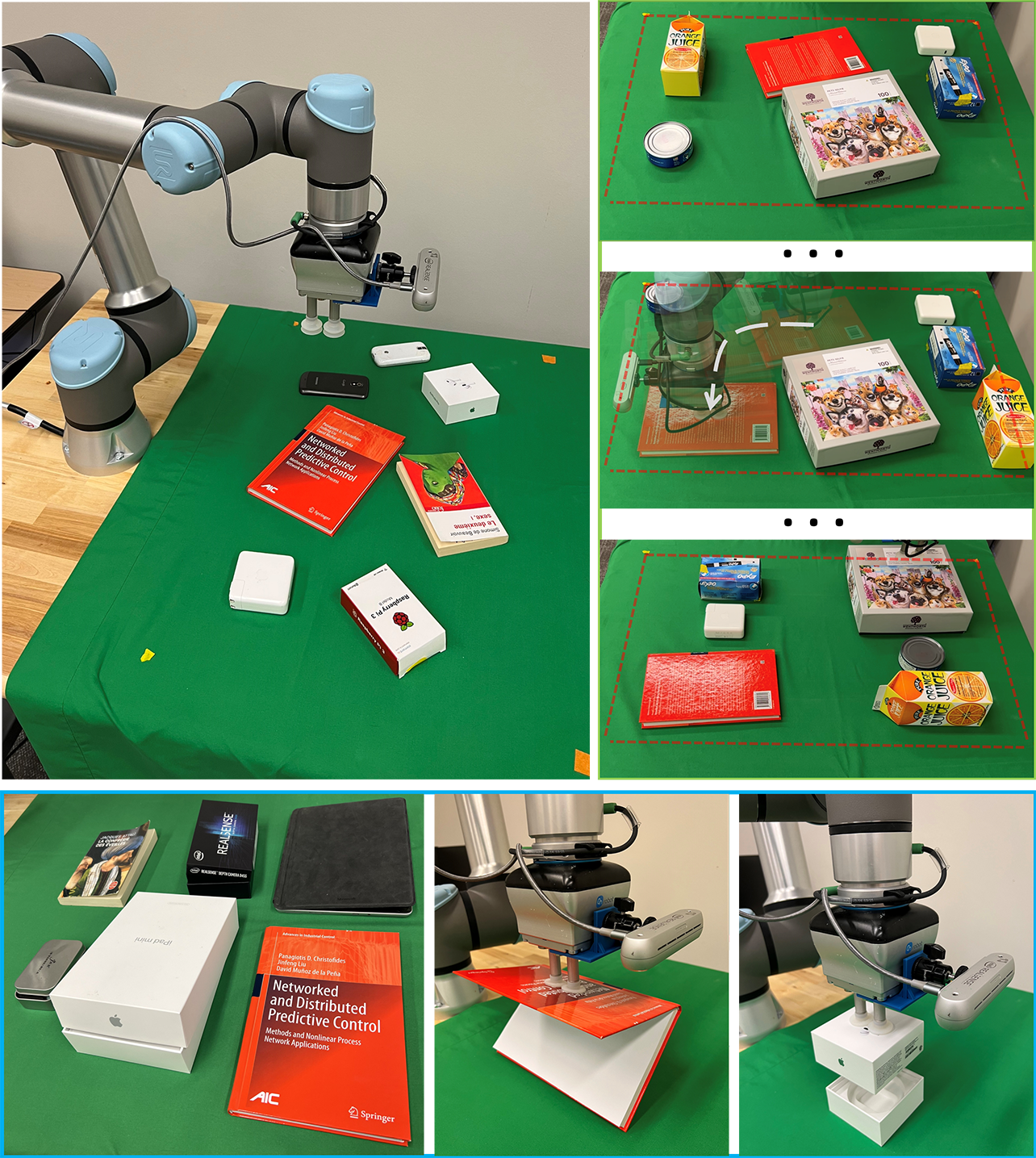}
        \put(1,33.5){\textcolor{white}{{\small (a) Hardware setup}}}
        \put(52,80){\textcolor{white}{{\small (b) Start state}}}
        \put(52,57){\textcolor{white}{\small (c) Push}}
        \put(52,33.5){\textcolor{white}{\small (d) Goal state}}
        \put(0.9,1){\textcolor{white}{\small (e) Objects require push}}
        \put(38,1){\textcolor{white}{{\small (f) Book}}}
        \put(64,1){\textcolor{white}{\small (g) Box}}
    \end{overpic}
    \caption{
    (a) Overview of system setup, a camera is mounted on the end-effector for perception.
    (b) - (d) An example case and an intermediate step in solving it.
    (e) Example objects requiring \emph{push}. (f:) \emph{pick-n-place} may break the book. (g) \emph{pick-n-place} will separate a box, failing to pick it up. 
    \label{fig:system-snapshot}
    }
\end{figure}

Toward quickly and optimally solving rearrangement tasks using multiple manipulation primitives, we focus on a tabletop setting where both \emph{pick-n-place} and \emph{pushing} are employed to rearrange objects (see Fig.~\ref{fig:system-snapshot}). Many objects, such as those shown in Fig.~\ref{fig:system-snapshot}(e), cannot be easily picked up and moved around without damaging or disassembling the object. For example, as shown in Fig.~\ref{fig:system-snapshot}(f)(g), books and certain boxes cannot be moved around using suction-based pick-n-place manipulation primitive (note that it is also difficult to do pick-n-place using fingered grippers). However, these objects can be effectively rearranged using a \emph{pushing} manipulation primitive in which the suction-based end-effector holds the object on or close to the tabletop and pushes/drags the object around (see Fig.~\ref{fig:system-snapshot}(c)). We call the frequently encountered yet largely unaddressed problem \emph{rearrangement with multiple manipulation primitives} (\prob). This study on \prob brings the following contributions: 

\begin{itemize}[leftmargin=4mm]
    \item With the formulation of \prob, we propose a first formal study of solving long-horizon rearrangement tasks utilizing multiple distinctive precision manipulation primitives with the goal of computing an optimized manipulation sequence. 
    Due to its high practical relevance, \prob constitutes an important specialized task and motion planning problem. 
    \item We developed two novel algorithms for \prob, the first of which is a fast rule-based solution capable of effectively and quickly solving non-trivial \prob instances. The second, leveraging Monte Carlo tree search (MCTS)~\cite{coulom2006efficient} and parallelism to look further into the planning horizon, delivers a much higher success rate for more challenging tasks, providing higher-quality solutions simultaneously. 
    \item We thoroughly evaluate our methods in simulation and extensive real robot experiments. In particular, our real robot experiments with integrated vision solutions, demonstrate that our algorithms can be readily applied to interact with everyday household objects in real-world scenarios.
\end{itemize}

\section{Related Work}\label{sec:related}
Robot manipulation \cite{mason2018toward} actions can typically be classified into two primary categories: prehensile~\cite{gao2022fast, wang2022efficient} and non-prehensile~\cite{zhang2023learning, yi2023precise, bai2022hierarchical}. 
In prehensile manipulation, a robot's gripper secures a grasp of the target object, allowing the object to move in with the end-effector. Fingered grasping and suction are two common prehensile manipulation actions. 
Non-prehensile manipulation leverages contact between objects and the environment, e.g., pushing, dragging, and toppling. 
While prehensile actions are limited in variety, non-prehensile actions abound \cite{mason2018toward,hou2020robust,doshi2022manipulation}. 
Limiting to tabletop rearrangement, pushing has been used extensively \cite{dengler2022learning, pezzato2023sampling} for relocating objects without lifting the object away from the surface.
However, pushing objects to follow a desired $SE(2)$ (i.e., 2D translation and rotation) trajectory with precision and speed is difficult. This leads to our design of a suction-based pushing manipulation primitive (Fig.~\ref{fig:system-snapshot}(c)), which is a mixed prehensile/non-prehensile primitive. 

Precisely and efficiently rearranging objects on a tabletop workspace, especially when object density is high, requires carefully allocating available space on the tabletop to facilitate the relocation of objects. 
%
%
Tang et al.~\cite{tang2023selective} consider removing objects from the workspace during the rearrangement of certain objects, reducing search space. We note that push is also used here but is used in an assistive capacity. 
Similarly, external space is used to temporally hold objects so that more space becomes free for rearranging in~\cite{gao2021running}.
Rearranging objects in confined dense clutter scenes is considered in~\cite{gao2022fast} where a heuristics-guided lazy search is applied.
In~\cite{lee2019efficient, ahn2022coordination, wang2022efficient}, rearrangement has been studied in shelve setups.
In prior research by Han et al.~\cite{han2018complexity}, the problem is translated into a graph-based formulation, factoring in the dependency graph among objects.
We note that these studies on tabletop arrangement exclusively work with pick-n-place manipulation. As such, these methods do not readily translate to \prob.

Many research efforts leverage reinforcement learning to address related challenges. 
The episodic nature of rearrangement often requires considering extended planning horizons. 
Consequently, combining MCTS with a predictive network is often an effective strategy. 
For instance, Huang et al.~\cite{huang2022interleaving} demonstrated the use of push actions to clear space, enabling the target object retrieval from cluttered environments. 
The concept of segregating objects into distinct regions using push actions is explored in~\cite{song2020multi}, while transitioning objects from initial to goal positions via push actions is discussed in~\cite{bai2022hierarchical}. 
Notably, all these studies employ predictive networks to accelerate the simulation phase of MCTS.
In contrast, this paper diverges from these approaches in that we primarily leverage MCTS for planning, which already demonstrates commendable performance. 
Even with recent advancements in network-based push prediction, as seen in~\cite{dengler2022learning}, there is still significant reliance on traditional path-planning algorithms for guidance. 
The efficacy of using predictive networks to direct MCTS, especially in push action scenarios necessitating object avoidance, remains an open question.

\section{Preliminaries}\label{sec:preliminaries}
\subsection{Rearrangement with Multiple Manipulation Primitives}
We now specify the concrete \emph{rearrangement with multiple
manipulation primitives} (\prob) studied in this work. 
Let the workspace be $\mathcal{W}$ is a 2D rectangle. 
The robot is provided with a start image (state) $s_s$ and a goal image (state) $s_g$ containing the initial and desired object arrangements.  
The robot must rearrange the objects to match the configurations specified in $s_g$. 
Two manipulation primitives are permitted: \emph{pick-n-place} $\mathcal{A}_{pp}$ and \emph{push} (from top) $\mathcal{A}_{pt}$. 
The robot's objective is to complete the task efficiently in terms of the \emph{execution time}. 
The start and goal states and the objects' transportation should be collision-free, and all objects should remain within the workspace. 
It is assumed that all tasks are feasible, i.e., there is always a viable solution $\mathcal{P} = \{a_1, a_2, ..., a_n\}$ leading from the start state $s_s$ to the goal state $s_g$, where $a \in \{\mathcal{A}_{pp}, \mathcal{A}_{pt}\}$.

A state $s_t$ represents the pose of objects at time $t$.
A pick-n-place action is specified by pick pose $(x_0, y_0, \theta_0)$ and place pose $(x_1, y_1, \theta_1)$.
A push action is specified by trajectories $\{(x_0, y_0, \theta_0), ..., (x_n, y_n, \theta_n)\}$, where the robot holds the object against the tabletop at the initial pose, and then push it following the waypoints, ending at the final pose. 


\subsection{Monte Carlo Tree Search}

The Monte Carlo tree search (MCTS) algorithm has broad applications. 
It is prevalent in turn-based tasks such as the game of Go~\cite{silver2016mastering}, but its usage extends beyond such contexts. 
MCTS plays a crucial role in solving rearrangement tasks~\cite{labbe2020monte, huang2022parallel, gao2023effectively}.
As an \emph{anytime} tree search algorithm, MCTS is designed to run for a fixed amount of time, each consisting of four stages: selection, expansion, simulation, and backpropagation. 
Fundamentally, MCTS preferentially exploits nodes that yield superior outcomes. 
To strike a balance between exploration and exploitation in the selection phrase, an \emph{upper confidence bound} (UCB)~\cite{auer2002finite} formula is utilized (see Eq.~\ref{eq:ucb1}), where $n$ is the parent node of $n'$, and $Q(n')$ is the total reward $n'$ received after $N(n')$ visits. 
\begin{equation}\label{eq:ucb1}
    \argmax\limits_{n' \in \text{children of } n} \frac{Q(n')}{N(n')} + c \sqrt{\frac{2\ln{N(n)}}{N(n')}}, 
\end{equation}

\vspace{0mm}
\section{Methodology}\label{sec:methods}
This work addresses the primary challenge of synergistic integration of pick-n-place and push actions. 
Because planning a push involves collision-free path planning in $SE(2)$, which is time-consuming, it poses a significant challenge if many push actions are explored. 
MCTS offers a solution capable of elegantly negotiating between the two disparate actions while maintaining optimality, given ample planning time. 

\subsection{Action Space Design}
Planning requires searching through candidate manipulation actions, which must first be \emph{sampled}. 
Action space design refers to action sampling, which plays a critical role in dictating the expansion of the tree search because there are an uncountably infinite number of possible manipulation actions. 
In sampling pick-n-place actions, we must ensure the place pose is collision-free. 
The same applies to a push action's final pose (though, in addition, the entire trajectory connecting all waypoints for a push action must be collision-free).
Four criteria are applied to sample the place/final poses for pick-n-place/push actions at the current state $s_t$: 
\begin{enumerate}[leftmargin=5mm]
  \item \textbf{Random}. A naive approach randomly samples collision-free place/final poses for pick-n-place/push actions.
  \item \textbf{Around Current and Goal}. 
  Random sampling is not always efficient. For object $o_i$, favoring regions around the current and goal poses of $o_i$ in $s_g$ can be helpful.
  \item \textbf{Grid}. Additionally, we adopt a grid-based sampling strategy. By modeling an object as a 2D polygon, we encapsulate it within a rotated bounding box. This allows us to generate a tiled representation in the workspace, denoted as $\mathcal{W}$, resembling a grid structure. This method proves advantageous for covering boundary areas, which can be hard to sample through random methods.
  \item \textbf{Direct to Goal}. If $o_i$ can be directly placed at its goal, this pose will be prioritized over the above three samplings in the search process. 
\end{enumerate}

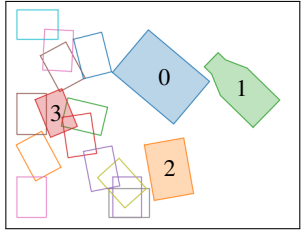
\begin{wrapfigure}[14]{r}{1.55in}
\vspace{-5mm}
    \makebox[\linewidth][c]{%
      \fbox{%
        \begin{overpic}[width=1.45in]{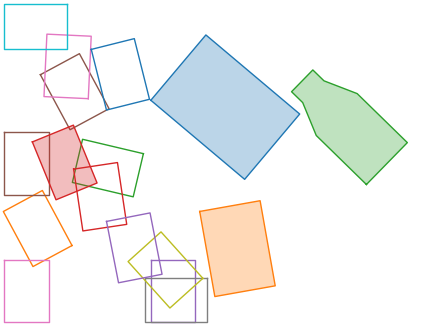}
            \put(52,49){\small 0}
            \put(80,45){\small 1}
            \put(54,16){\small 2}
            \put(13,36){\small 3}
        \end{overpic}%
      }%
    }
    \caption{Consider action sampling for labeled $3$ to be manipulated using push (there are a total of four objects). The absence of sampled actions in the right region is attributed to obstructions posed by objects $0$, $1$, and $2$, preventing the movement of object $3$ to that area.
    \label{fig:sample-actions}
    }
\end{wrapfigure}

Once the place/final poses have been sampled, a $\mathcal{A}_{pp}$ sample is obtained.
However, $\mathcal{A}_{pt}$ requires an additional step - generating a trajectory from the current pose to the place pose for $o_i$.
We employ RRT-connect~\cite{844730} during the tree search and LazyRRT~\cite{bohlin2001randomized} for robot execution to produce such collision-free trajectories with a set time limit. 
An example of sampling the final pose for a push action is shown in Fig.~\ref{fig:sample-actions}.
The criteria for sampling are crucial in solving the problem. Relying solely on standard uniform sampling often proves inefficient, particularly when sampling around boundaries and certain edge cases. While one might consider increasing the sample size to cover these edge cases, this inadvertently leads to many redundant actions that are time-consuming to process.

\subsection{Hierarchical Best-First Search}
The first algorithm we designed for \prob is a (greedy) best-first search algorithm called \emph{hierarchical best-first search} (\hbfs). \hbfs is outlined in Alg.~\ref{alg:hbf} and operates according to the following sequence of steps: 
\begin{itemize}[leftmargin=4mm]
    \item (Lines 3-4) When objects can be directly moved to their goal poses, an action cost is computed for each. The action yielding the smallest cost is then applied.
    \item (Lines 5-9) For each object $o_i$, \hbfs identifies which objects occupy $o_i$'s goal and attempts to displace these obstructing objects in the direction of their respective goals. If no actions are feasible in this direction, a random action is sampled. Again, the action associated with the smallest cost is selected and implemented.
    \item (Line 10) If the above steps do not yield a viable action, an action is randomly selected for execution.
\end{itemize}
The above three phases of \hbfs may best be viewed as a three-level hierarchical search. To boost its performance and solution optimality, \hbfs is implemented by leveraging multi-core capabilities of modern CPUs. This is realized by executing multiple \hbfs in parallel and choosing the best action among the returned solutions.

\begin{algorithm}
    \begin{small}
    \DontPrintSemicolon
    \SetKwFunction{FHBF}{HBFS}
    \Fn{\FHBF{$s_s, s_g$}}{
        $s \gets s_s$, $A \gets \varnothing$ \;
        \lFor{$o_i$ \text{in} $s$}{
            $A \gets A \cup \{\text{move } o_i \text{ to its goal}\}$
        }
        \lIf{$|A| > 0$}{
            \Return the lowest cost action from $A$
        }

        \For{$o_i$ \text{in} $s$}{
            $obs \gets \text{objects occupy the goal pose of } o_i$ \;
            \For{$o_j$ \text{in} $obs$}{
                $A \gets A \cup \{\text{move } o_j \text{ towards its goal, otherwise at random}\}$
            }
        }
        \lIf{$|A| > 0$}{
            \Return the lowest cost action from $A$
        }

        \Return a randomly sampled action
    }
    \caption{Hierarchical Best-First Search (\hbfs)}\label{alg:hbf}
    \end{small}
\end{algorithm}

\subsection{Speeding up MCTS with Parallelism}
Standard MCTS is more straightforward to implement, but it does not fully utilize multi-core processing capabilities of modern hardware. 
We introduce parallelism to the expansion and simulation stages of MCTS, leverage tree parallelization techniques~\cite{chaslot2008parallel}. 
The application of parallelism allows for decoupling the select and expand stages from the simulation stage in an MCTS iteration. 
The decoupling allows multiple MCTS iterations to be carried out simultaneously, limited only by the number of CPU cores. 
The standard UCB formula used in MCTS is updated as Eq.~\ref{eq:ucb2},
\begin{equation}\label{eq:ucb2}
    \argmax\limits_{n' \in \text{children of } n} \frac{Q(n')}{N(n') + \hat{N}(n')} + c \sqrt{\frac{2\ln{(N(n) + \hat{N}(n))}}{N(n') +\hat{N}(n')}},
\end{equation}
where the idea of virtual loss~\cite{chaslot2008parallel} is applied by adding one extra virtual visit counts $\hat{N}$ that indicates a node has been selected but not yet simulated and backpropagated.
Since the simulation and subsequent backpropagation stages are not yet completed, the tree search algorithm must be notified to update the $Q$ and $N$ of the node. This adjustment minimizes the likelihood of revisiting the node in the next iteration, implementing a conservative approach in anticipation of a potentially poor reward. Once the simulation stage has concluded, $Q$ is updated in backpropagation with returned reward from simulation result, $N$ increments and $\hat{N}$ decrements.

\subsection{Adaption MCTS for \prob}
Given \prob's extremely large search space due to push actions' trajectory planning requirements, modifications are introduced to best apply MCTS to \prob. 

\textbf{Action Space Bias}. The action space used in the simulation stage is a subset of that used in the expansion stage. 
Given that one iteration of the simulation stage constitutes a coarse estimation of action and state, reducing the number of actions leads to a larger number of total iterations.

\textbf{Biased Simulation}. A straightforward implementation of the simulation stage in MCTS is a random policy that indiscriminately selects an action for execution, ultimately obtaining a reward at the final state reached. 
In our implementation, we adopt a heuristic to guide the action selection in the simulation stage towards the ultimate goal of a given object. Specifically, with a probability of $\theta_{sim}$ (dynamically changed based on depth of the search), a random action is selected; otherwise, an attempt is made to select an action that will put an object on its goal pose. 
This introduces a bias in the simulation stage, which offsets the drawback of limited iterations due to the time-consuming nature of motion planning and collision checking.
We note that we did not adopt a recent advancement in MCTS for long-horizon planning that injects a data-driven element to partially learn the reward, e.g., a neural network can be trained to evaluate the quality of an action-state pair~\cite{song2020multi, bai2022hierarchical, huang2022interleaving}.

\textbf{Reward Shaping}. We structure the reward to favor the goal state but without introducing undue bias. The reward function plays a critical role as it steers the tree search, composed of three components.
Firstly, if the task is accomplished, with all objects placed at their goal poses, a reward of $R_g$ is awarded.
Secondly, if an object $o_i$ is located at its goal pose, a reward $r_o$ is given. The cumulative reward from all objects, denoted as $R_o$, is computed as $R_o = \sum_{i} r_{o_i}$. 
Lastly, the reward structure also takes into account the cost associated with the movement of objects.
For the pick-n-place action ($\mathcal{A}_{pp}$), the cost corresponds to the Euclidean distance between the pick and place poses, with an additional fixed cost factored in. 
For the push action ($\mathcal{A}_{pt}$), the cost is determined by the Euclidean distance of the path, also supplemented by a fixed cost.
Additionally, a base reward is computed $R_b = R_o(s_0)$ from initial state $s_0$, which is used to normalize the final reward during the search.
For each iteration, a reward is returned by the simulation stage and is updated during the backpropagation stage.
\[
    R_i= 
\begin{cases}
    \max(0, R_g - \textit{cost} - R_b),& \text{if } s_i \text{ is goal state}\\
    \max(0, R_o(s_i) - \textit{cost} - R_b),              & \text{otherwise}
\end{cases}
\]

Traditionally, $Q$ retains the average reward values derived from simulation results, providing a robust estimation for action over millions of iterations. 
However, in our case, we aimed to maintain the planning time within reasonable limits. 
Therefore, we introduced a priority queue to store simulation results, which serves as the $Q$ value in the algorithm.
For the purpose of calculation in Eq.~\ref{eq:ucb2}, we only preserve the top $k$ rewards, similarly in the previous work~\cite{huang2022interleaving}.
There is a possibility that during a simulation, a subsequent action may transition the state to one with lower rewards, thereby negating the benefits of a preceding beneficial action within that simulation. 
To limit the search time, we choose to return the maximum reward encountered at the intermediate steps during the simulation instead of the final reward. 
Therefore, the returned reward from a simulation is given by $$\max(\beta \cdot \max_{i \in m-1}(R_i), R_m) \cdot \gamma^m,$$ where $\beta\in (0,1]$ is a scaling parameter and $m$ is the total steps used in the simulation.
$\gamma$ is the discount factor, encouraging the problem to be solved in the early stage if possible.

Due to limited space, we omit the pseudo-code of the parallel MCTS algorithm but note that all details for reproducing the algorithm have been fully specified. We call the resulting algorithm \emph{parallel Monte Carlo tree search for multi-primitive rearrangement} or \pmmr.

\section{Experimental Evaluation}\label{sec:experiments}
We evaluate \hbfs and \pmmr methods for \prob in simulated environments and on a real robot. 
Regardless of whether it is a simulation or a real robot experiment, both algorithms perform \emph{percept-plan-act} loops until the task is solved or the budgeted time is over. 
Extensive ablation studies are conducted to scrutinize different design choices.
All experiments were conducted on an Intel i9-10900K (10 CPU cores) desktop PC and run in Python.
As a note, limited testing shows that using Intel i9-13900K (24 CPU cores) reduces the planning time by roughly half, demonstrating the effectiveness and scalability of employing parallelism. 

\begin{figure}[h]
    \centering
    \includegraphics[width = 0.99\linewidth]{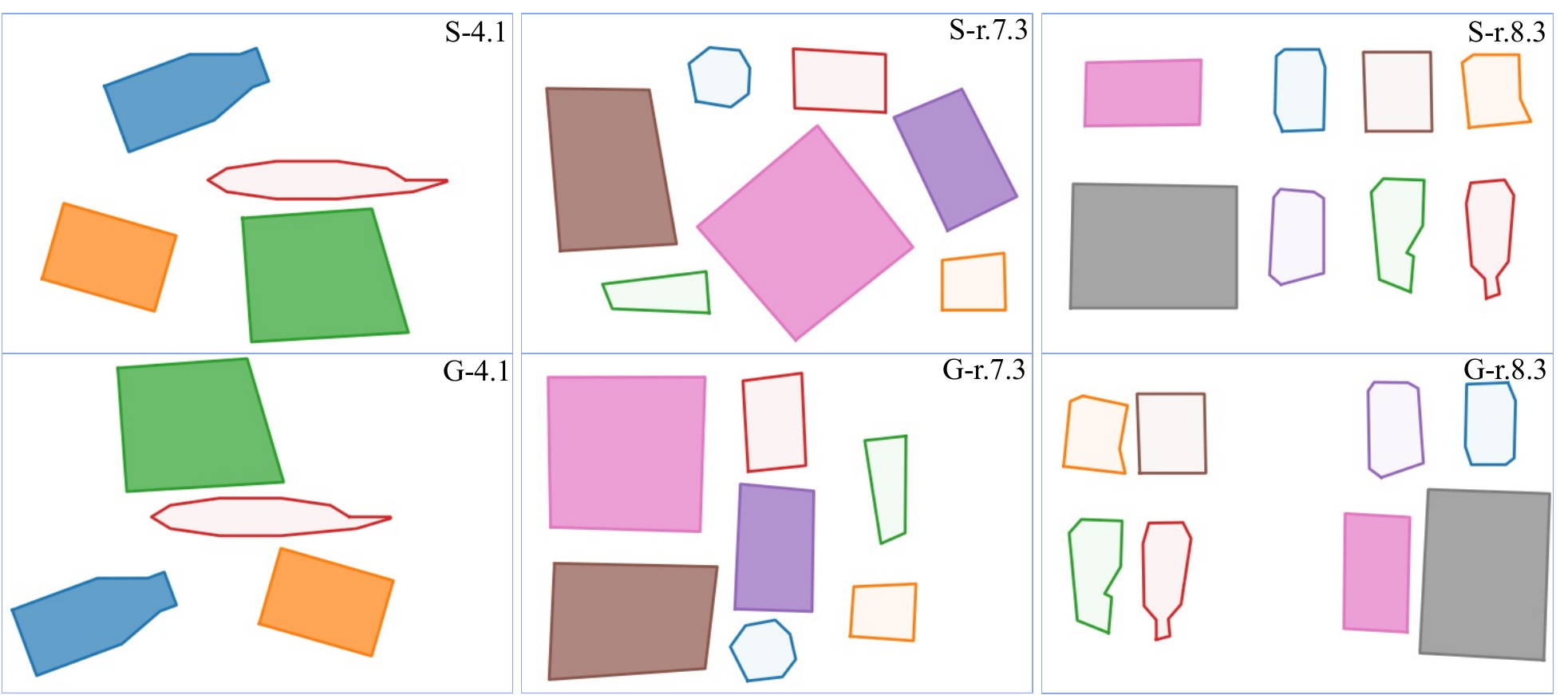}
    \caption{Example cases. The top row shows the start states and the bottom goal states. Lightly shaded objects can be pick-n-placed; heavily shaded objects must be manipulated using push. %
    Cases 4.1, r.7.3, and r.8.3 are evaluated and presented in Fig.~\ref{fig:sim}. Objects are distinguished by color.
    \label{fig:cases}
    }
\end{figure}

\subsection{Simulation Studies}
Simulation is conducted in PyBullet~\cite{coumans2016pybullet}. 
A real robot setup, consisting of a Universal Robot UR5e $+$ OnRobot VGC-10 vacuum gripper, is replicated. 
The robot operates under end-effector position control; the workspace measures  $0.78 \times 0.52 \si[per-mode=symbol]{\meter\squared}$.
$25$ feasible scenarios are created where all objects are confined within the workspace. Object sizes, shapes, and poses are randomly determined in each scenario (see Fig.~\ref{fig:cases} for some examples). 
Cases that are trivial to solve (e.g., objects that happen to be mostly small) are filtered. 
The number of objects ranges from four to eight; five distinct cases are generated for each specified number of objects.

\begin{figure*}[ht!]
\vspace{1mm}
    \centering
    \begin{minipage}{\linewidth}
        \includegraphics[width = .99\linewidth, height=1.35in]{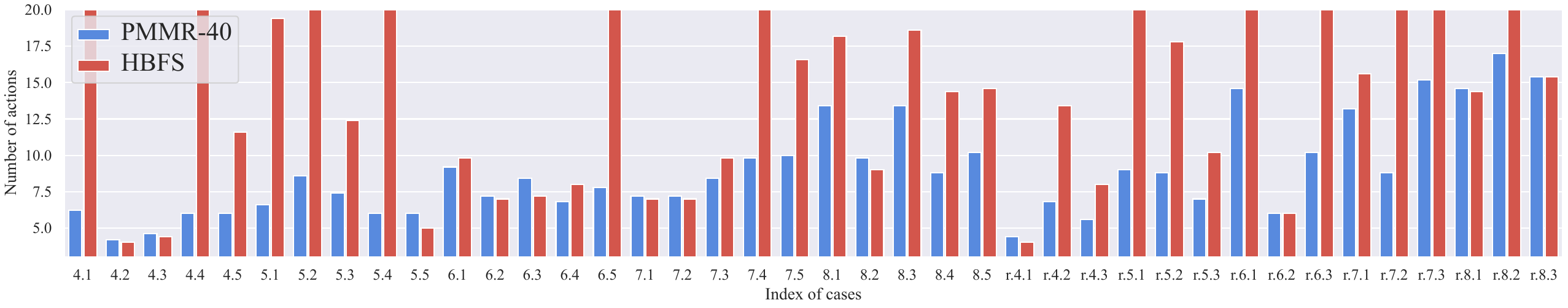}
    \end{minipage}
    \vspace{1mm}
    \begin{minipage}{\linewidth}
        \includegraphics[width = .99\linewidth, height=1.35in]{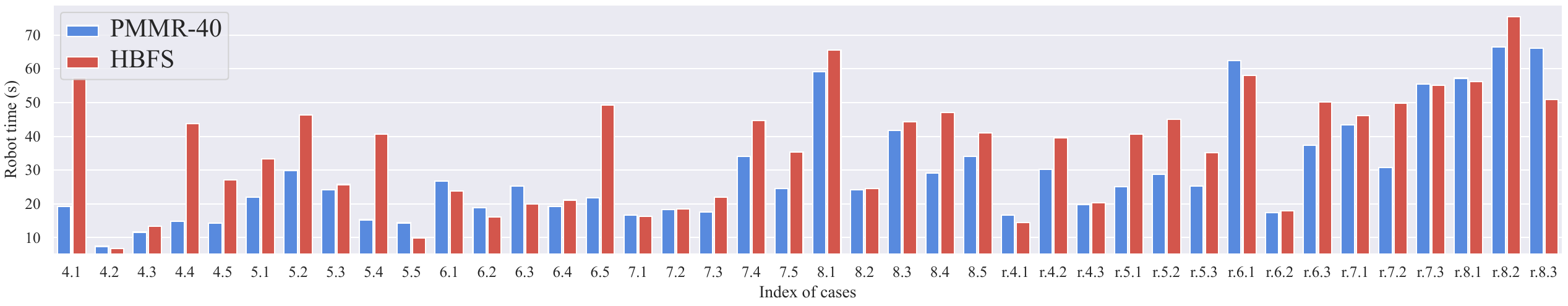}
    \end{minipage}
    \vspace{-1mm}
    \caption{\label{fig:sim}
        As an expanded illustration of Tab.~\ref{tab:simtable}, the upper plot lists the number of actions the robot executes to resolve individual cases. 
        The lower plot lists the robot's execution times in solving the individual cases following the computed plan.
        For the labels on the horizontal axis, the first digit indicates the number of objects contained within each case, while the second digit represents the index of the cases. Cases beginning with the prefix 'r' are the ones that are constructed for and executed by the real robot setup.
    }
\vspace{-2mm}
\end{figure*}

In evaluating \pmmr, each \emph{percept-plan-act} loop runs for a predetermined duration to identify the best next action until the problem is resolved or the maximum number of actions has been exhausted. 
If the latter occurs, it is treated as a failed case. 
%
We denote the corresponding \pmmr method as \pmmr-X, where X is the maximum number of seconds allowed in a single iteration of the loop.
We settled on \pmmr-40 as the main \pmmr method used in the evaluation. 
In both simulation and real robot experiments, for \pmmr-40, we keep the top $k=100$ for $Q$ value, $c=1.5$ in Eq.~\ref{eq:ucb2}.
The max MCTS tree depth $D$ is based on the number of objects $\mathcal{N}$: $D=2\mathcal{N} + 2$.
$\theta_{sim}$ is based on the depth $d$ of the node $\theta_{sim} = \max(-0.106 + 0.231d - 0.013d^2, 0.2)$.
$r_o=0.7$ for object can be operated by $\mathcal{A}_{pp}$, the $R_g=2r_0 \mathcal{N}$. For an object that can be operated by $\mathcal{A}_{pt}$, the reward is given to $1.1r_o$.
We set $\beta=0.5$ and $\gamma=0.9$ to scale the reward.

\begin{table}[h]
    \centering
    \scalebox{0.99}{\begin{tabular}{c|c|c|c|c}
        & Robot Time & Completion & Num. of Actions & Plan Time     \\ \hline
        \pmmr-40 & $29.17$s & $98.00\%$ & $8.90$ & $264.99$s \\ \hline
        \hbfs & $36.22$s & $54.50\%$ & $13.72$ & $30.04$s \\ \hline
    \end{tabular}}
    \caption{Simulation experiment results for 25 simulated cases and 15 real robot experiments for \hbfs and \pmmr-40.}
    \label{tab:simtable}
\end{table}

Individual experiment results for all cases are shown in Fig.~\ref{fig:sim} (which also includes cases used for real-robot experiments, to be detailed in Sec.~\ref{sec:real}).
Detailed experiment results are presented in Tab.~\ref{tab:simtable}.
Here, \emph{robot time} refers to the cumulative time required for the robot to execute all actions, while \emph{completion} refers to the success rate. The number of actions quantifies the execution of atomic actions, represented by $\mathcal{A}_{pp}, \mathcal{A}_{pt}$. The \emph{plan time} is the total planning time.
Each case underwent five independent trials.

Failure often happens because the case requires more than 15 actions to solve (even for humans). A failure may be due to sampled actions not containing a solution or the search not being deep enough. Sometimes, the algorithm can recover from an early bad choice, but not always since the number of iterations is capped. 
We observe that, while \hbfs runs relatively fast in comparison to \pmmr-40, it frequently fails ($55\%$ success v.s. $98\%$ for \pmmr-40) and uses many more actions (13.7 v.s. 8.9 for \pmmr-40). 
Visually, as can be seen in the accompanying video, the actions generated by \pmmr-40 are much more human-like than those by \hbfs (same holds for real robot experiments).

\subsection{Ablation Studies}
\begin{figure}[h]
    \centering
    \includegraphics[width = \linewidth]{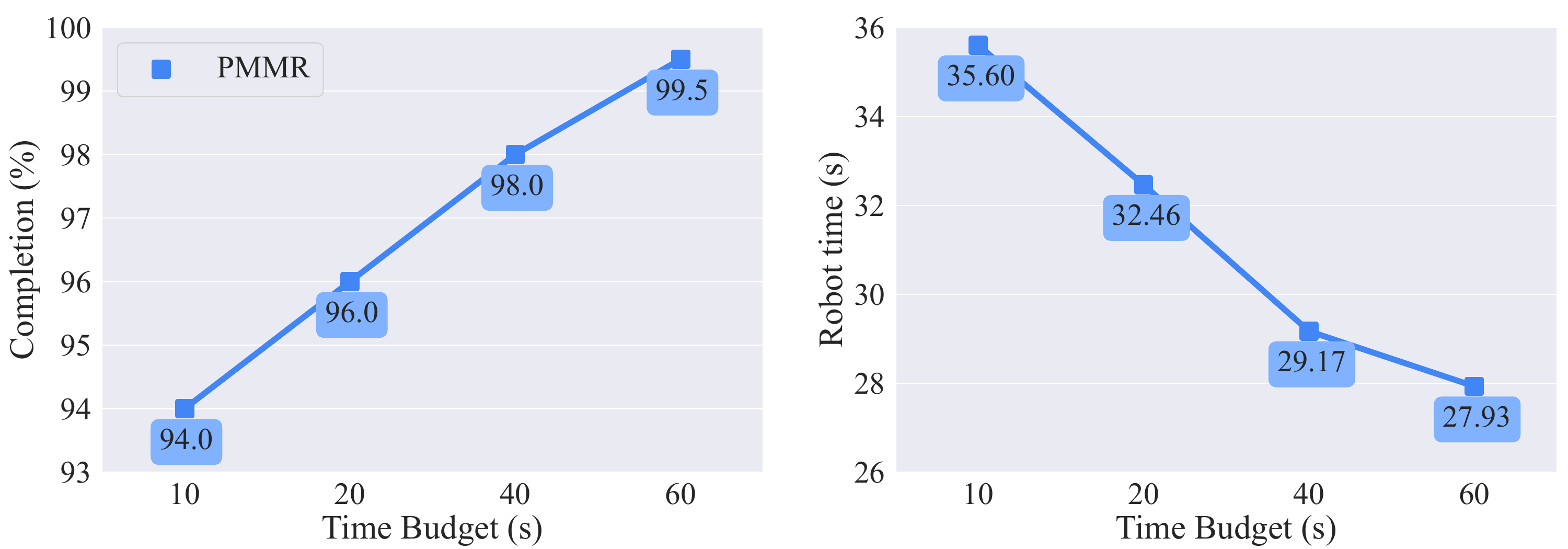}
    \caption{\label{fig:simablation}
        \pmmr is evaluated with different time budgets. The reported values are averaged over 40 cases.
    } 
\end{figure}

We investigated the impact of time budgets, the depth of tree search, and the base reward $R_b$ on solving \prob. The time budget is critical; an extended search duration tends to yield better results. However, it is necessary to balance planning time and solution quality. An insufficient search might sample highly suboptimal paths, leading to locally optimal actions. As depicted in Fig.~\ref{fig:simablation} and Tab.~\ref{tab:ablation-table} shows the correlation between planning time and solution quality, leading us to select \pmmr-40 for our main evaluation.

A shallow MCTS (\pmmr-40 ($D=3$), max MCTS tree depth of $3$) was included specifically to compare with \hbfs, which has three ``depth levels'' per iteration. 

In terms of reward design, as detailed in section~\ref{sec:methods}, we introduced a base reward $R_b$, which serves to normalize the reward to 0 at root. Without this adjustment, the tree search might commence with a non-zero reward signal, where a disadvantageous branch may still return a reward, causing the search to frequently explore such branches. By comparing \pmmr-40 (no-$R_b$) in Tab.~\ref{tab:ablation-table} with \pmmr-40 in Tab.~\ref{tab:simtable}, we observe that the introduction of $R_b$ aids the tree search.

\begin{table}[h]
\vspace{2mm}
    \centering
    \scalebox{0.94}{\begin{tabular}{c|c|c|c|c}
        & Robot Time & Completion & Num. of Actions & Plan Time     \\ \hline
        \pmmr-10 & $35.60$s & $94.00\%$ & $10.51$ & $103.61$s \\ \hline
        \pmmr-20 & $32.46$s & $96.00\%$ & $9.86$ & $155.68$s \\ \hline
        \pmmr-60 & $27.93$s & $99.50\%$ & $8.53$ & $358.44$s \\ \hline
        \pmmr-40 ($D=3$)& $57.83$s & $32.50\%$ & $16.62$ & $641.76$s \\ \hline
        \pmmr-40 (no-$R_b$)& $32.92$s & $94.00\%$ & $9.83$ & $270.65$s \\ \hline
    \end{tabular}}
    \caption{Ablation studies for all $40$ cases to be compared to Tab.~\ref{tab:simtable}.}
    \label{tab:ablation-table}
\end{table}

\subsection{Real Robot Experiments}\label{sec:real}

In simulation, we emulate the vacuum function by attaching the object to the end-effector using an extra link via a fixed joint. 
We employ two vacuum cups to provide sufficient suction power to ensure a robust connection in the real-world setup. 
The point of suction on the object is taken as its center, assuming this central area is flat.
Similar to simulation studies, the number of objects ranges from four to eight, and three distinct cases are generated for each number of objects.

A RealSense D455 camera is affixed to the robot's wrist, capturing the scene from a top-down perspective, and an orthogonal view is rendered from the point cloud.
We employ the Segment Anything Model~\cite{kirillov2023segany} to extract masks of the objects present on the table. 
Subsequently, OpenCV~\cite{opencv_library} is applied to determine the contours and approximate them into polygons, which are then used for planning.
An additional step determines each object's $SE(2)$ poses. 
For each case, the experiment is repeated at least three times.

\begin{figure}[h]
    \centering
    \includegraphics[width=\linewidth, height=2.45in]{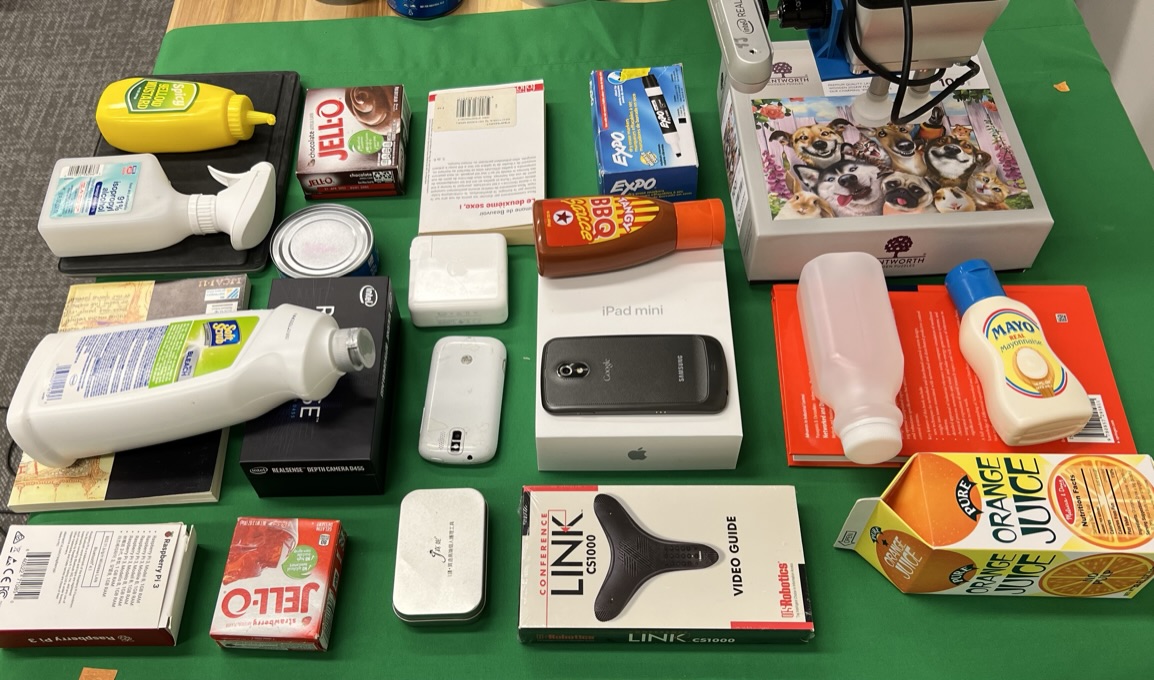}
    \caption{The full set of objects used in our real robot experiments. 
    \label{fig:all-objects}
    }
\end{figure}

\begin{figure}[ht!]
    \centering
    \includegraphics[width =\linewidth, height=1.35in]{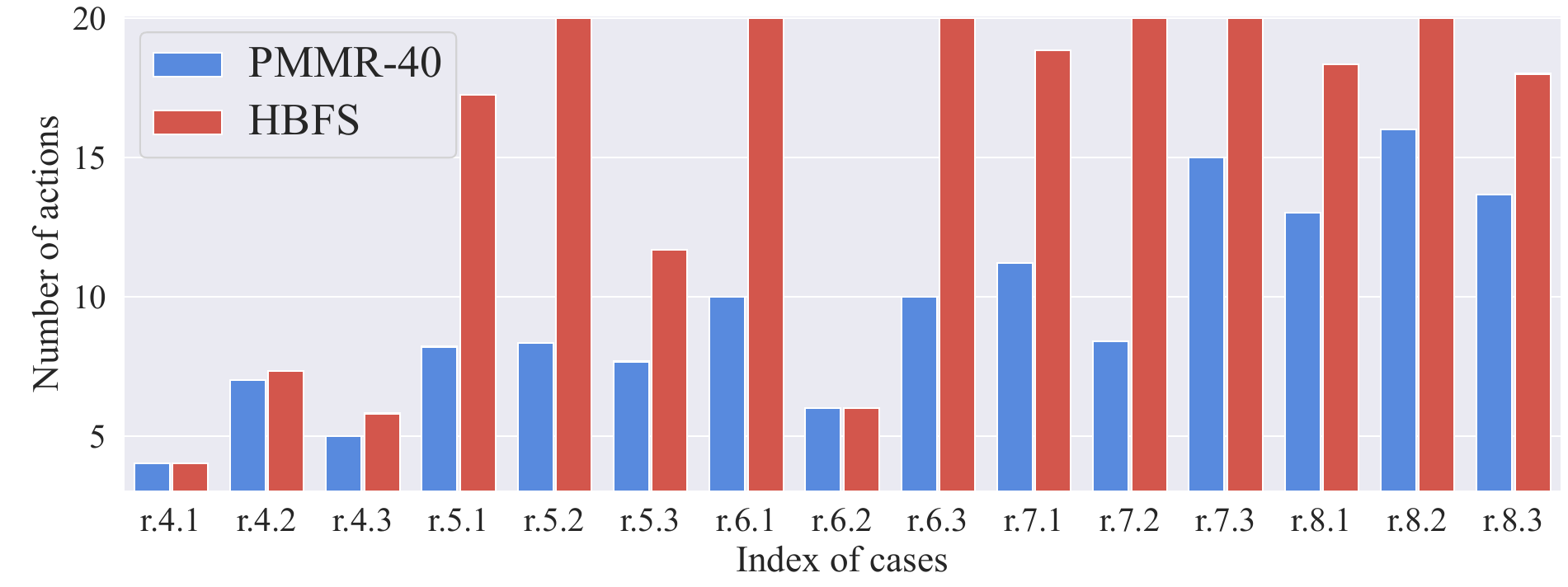}
    \caption{\label{fig:realnum}
        As an expanded illustration of Tab.~\ref{tab:realtable}, this plot illustrates the number of actions the robot executes to resolve individual cases. 
    } 
\vspace{-2mm}    
\end{figure}

Due to the small sim-to-real gap, We let the algorithm plan the entire sequence of manipulation actions at the beginning, which generally works well. 
If no solution is found using $20$ actions, we mark it as a failure; otherwise, the robot executes the actions. 
Robot time is not recorded for failure cases, hence its absence in Tab.~\ref{tab:realtable}. 
For completeness, in cases where both \pmmr and \hbfs succeed at least once, \hbfs averages $15.15$ actions and \pmmr $9.56$, with robot (execution) times of $96.62$ seconds and $95.75$ seconds, respectively (but, note that \hbfs fails much more frequently).
Results from individual benchmarks across 15 cases are presented in Fig.~\ref{fig:realnum}. Each case was subjected to three independent trials.
In real robot experiments, the cases are intentionally designed to be challenging to solve. 
A greedy action may exacerbate the problem, making it even more difficult to resolve. 
Consequently, the completion rate of \hbfs is significantly reduced.

\begin{table}[h]
    \centering
    \scalebox{0.96}{\begin{tabular}{c|c|c|c|c}
        & Robot Time & Completion & Num. of Actions & Plan Time     \\ \hline
        \pmmr-40 & $95.75$s* & $96.44\%$ & $9.56$ & $292.02$s \\ \hline
        \hbfs & $96.62$s* & $38.33\%$ & $15.15$ & $29.05$s \\ \hline
        \pmmr-40 (Sim) & $-$ & $94.67\%$ & $10.44$ & $306.41$s \\ \hline
        \hbfs (Sim) & $-$ & $45.33\%$ & $14.99$ & $22.18$s \\ \hline
    \end{tabular}}
    \vspace{2mm}    
    \caption{Experiment results of real robot trials across 15 cases, with time budgets constrained to a maximum of 40 seconds for a single MCTS run. The robot time is only considered in cases where both methods succeed at least once. Additionally, benchmarks from simulations covering 15 cases are included for sim-to-real gap comparisons. The robot time for \pmmr-40 and HBF, denoted with an asterisk, is recorded only for successful cases.}
    \label{tab:realtable}
\vspace{-4mm}    
\end{table}

\section{Conclusion and Future Directions}
We make the observation that humans frequently solve manipulation challenges using multiple types of manipulation actions. In contrast, there has been relatively limited research tackling planning high-quality resolutions for long-horizon manipulation tasks exploring the synergy of multiple manipulation actions. 
Inspired by how humans solve everyday manipulation tasks, in this paper, we propose and study the \emph{rearrangement with multiple manipulation primitives} (\prob) problem. 
Toward optimally solving \prob, we developed two effective methods, \hbfs and \pmmr, with \pmmr especially adept at solving difficult \prob instances with high success rates and producing high-quality solution sequences, which are confirmed with thorough simulation and real robot experiments, going through the full percept-plan-act loops. 

With the current work paving the way, in future studies, we plan to expand the research in two directions: (1) We will explore adding additional manipulation actions (e.g., pushing multiple objects, flipping objects) and solving more difficult rearrangement tasks (e.g., stacking to form complex structures), and (2) Leveraging the current algorithms to generate training data, we will develop data-driven methods (e.g., machine learning and/or reinforcement learning) to further speed up the planning process. 

\bibliographystyle{IEEETran}
\bibliography{references}

\end{document}